\documentclass[a4paper]{cas-sc}

\usepackage[numbers]{natbib}
\usepackage{algorithm,algpseudocode}
\usepackage{program}
\usepackage{amsmath}
\usepackage{tabularx}
\usepackage{booktabs}
\usepackage{graphicx}
\usepackage{grffile}
\usepackage{float}
\usepackage{textcomp}
\usepackage{subcaption}
\usepackage{placeins}
\usepackage{caption}
\usepackage{enumitem}
\usepackage{float}
\newlist{steps}{enumerate}{1}
\setlist[steps, 1]{label = Step \arabic*:}

\begin{document}
\title [mode = title]{Fine-Grained Emotion Detection on GoEmotions: Experimental Comparison of Classical Machine Learning, BiLSTM, and Transformer Models}  
\shorttitle{Sentiment Analysis on GoEmotion Dataset}
\shortauthors{Ani Harutyunyan}

%
\author[]{Ani Harutyunyan}
\ead{<ani_harutyunyan4@edu.aua.am>}

\author[]{Sachin Kumar}
\cormark[1]

\ead{<s.kumar@aua.am>}
\affiliation[]{organization={Zaven P. and Sonia Akian College of Science and Engineering},
         addressline={American University of Armenia}, 
          city={Yerevan},
            postcode={0019}, 
           country={Armenia}}
           
\cortext[1]{Corresponding author}
\fntext[1]{}


\begin{abstract}
Fine-grained emotion recognition is a challenging multi-label NLP task due to label overlap and class imbalance. In this work, we benchmark three modeling families on the GoEmotions dataset: a TF-IDF-based logistic regression system trained with binary relevance, a BiLSTM with attention, and a BERT model fine-tuned for multi-label classification. Experiments follow the official train/validation/test split, and imbalance is mitigated using inverse-frequency class weights. Across several metrics, namely Micro-F1, Macro-F1, Hamming Loss, and Subset Accuracy, we observe that logistic regression attains the highest Micro-F1 of 0.51, while BERT achieves the best overall balance surpassing the official paper's reported results, reaching Macro-F1 0.49, Hamming Loss 0.036, and Subset Accuracy 0.36. This suggests that frequent emotions often rely on surface lexical cues, whereas contextual representations improve performance on rarer emotions and more ambiguous examples.
\end{abstract}




\maketitle
\section{Introduction}
Understanding emotions in text is extremely important for the development of artificial intelligence. This matters because of the wide variety of applications it can have in everyday life, such as conversational agents, customer feedback, emotion aware moderation, etc. A more powerful skill is to understand fine-grained emotion types and not just the usual 7-8 emotions. GoEmotions dataset \cite{r4} is a perfect example for such case, which has 28 different emotions. This dataset is particularly challenging for number of reasons, such as that it moves beyond the binary positive and negative classification, providing fine grained emotion labels. Each sentence has one or more labels, so cooccurence is possible and we have extreme class imbalance. In this paper, a multilabel classification has been done on this dataset comparing three different models with increasing capacities. Starting with baseline model, then adding neural nets, and finalizing with transformer architecture. The models are TF-IDF + logistic regression using binary relevance, a BiLSTM with attention, and  fine-tuned BERT. These models were evaluated with a number of metrics to understand how they work as deeply as possible.


Contribution of the paper
\begin{itemize}
    \item This work implements and compares three model families (Logistic Regression, BiLSTM+Attention, and fine-tuned BERT) on the GoEmotions dataset under the same experimental setup.
    \item All models are evaluated using various multi-label metrics to provide a comprehensive comparison, such as Micro/Macro F1, Hamming Loss, Subset Accuracy, etc.
    \item An analysis is done discussing how class imbalance and per-emotion frequency affect performance, including the difference between common versus rare emotion labels.
\end{itemize}

The rest of the paper is organized as follows: Section 2 describes the Background and Related Work; Section 3 presents the Methodology with Section 3.1 describing the dataset and giving some statistics, Section 3.2 explaining the Proposed Framework and Section 3.3 showing the Evaluation Measures/Metrics;  Section 4 reports the Experiments, Results and Discussion; Section 5  covers the Conclusion, Limitations and Future Directions.

\section{Background and Related Work}
Emotion recognition has been a popular field of interest for many years in NLP tasks, however, there was the problem of good quality dataset: most of them were either automatically labeled or had significantly small samples in the dataset. Thus, there were no famous benchmarks for the state of the art models available. GoEmotions dataset \cite{r4} provided a new valuable contribution to the research done on emotion datasets, becoming the new benchmark in this field. However, the task was quite complicated, 27 emotions + 1 neutral emotion, severe class imbalance and human subjectivity. Different models have been built and a lot of work has been done for this classification problem, with both classical machine learning and deep learning models. The authors have reported the results of evaluations with BERT. In the paper called \textit{"Fine-Grained Emotion Prediction by Modeling Emotion Definitions" } \cite{r11} Singh et al have introduced a multi-task architecture which was using both emotion labels and also their definitions with semantic ideas and meanings. With this paper, they showed that with external knowledge it was possible to generalize on GoEmotions dataset. In our paper, in contrast to their proposed model, we do not use external knowledge, and our models are solely trained on the dataset itself, which makes the problem more difficult. Another paper on GoEmotions dataset, tries to address the scarcity of some labels by identifying the rare emotions \cite{r8}. They finetune BERT with different loss functions to find the best and suitable one for such imbalanced dataset. Their findings help us to understand the low performance of BiLSTMs and other simpler models. Wang et al. \cite{r15} try to go deeper into LLMs and their performance on this dataset, they improve BERT finetuning with data augmentation and other techniques, however, their experiments require much more computational power than ours, and therefore, when comparing the performance, the resources should be taken into account as well. In their paper called \textit{"Comparative Study of Machine Learning and Deep Learning Techniques"} \cite{r1}, the authors compare classical ML models with deep neural nets and report results of SVM, LSTM and other neural model performances, which closely relate to our motivation of training and evaluating three models - Logistic Regression with Binary Relevance, BiLSTM with Attention block and BERT with multi-label head. Overall, in spite of the fact that a number of studies have been done in Emotional Recognition and Classification using machine learning models, recurrent neural networks, transformer based models and even large language models, most of the papers have done the experiments with different configurational setups or have compared the results on differing datasets. Thus, there are limited experiments done to clearly and directly compare the results of different capacity models on the GoEmotions dataset. To address this limitation, in our work we compare Logistic Regression with Binary Relevance \cite{r5, r12}, BiLSTM with Attention \cite{r2, r6, r10, r13} and BERT with multi-label head \cite{r3, r9, r12} on fine-grained, multi-label emotion classification task with GoEmotions dataset.
\section{Methodology}
This section describes the overall methodological approach that was followed for this study. It includes details on resources used, the design of the proposed system, and the criteria applied in assessing the performance of the system. Together, these elements form the basis for how the experiments were conducted and how the results should be interpreted.

\subsection{Dataset Description}

The dataset used in this study is the GoEmotions dataset \cite{r4}, a large-scale, manually annotated corpus designed for fine-grained emotion classification. It was released by Google Research in 2020.  It provides 28 different fine-grained emotions, going beyond the usual binary or ternary sentiment categories. It is the largest manually labeled high quality data in this category. The dataset is derived from English Reddit comments made within the the period 2005-2019, a large social media platform where different types of emotions and characteristics meet each other. The choice of the dataset can be explained with different reasons. First of all, Reddit comments represent authentic emotions because they are not polished formal sentences; instead, they show people's true perspectives, thoughts and opinions written in slang, and usually with emojies. Reddit covers the most different domains starting from finances up to fashion, which makes the comments very diverse. Last but not least, the wide temporal range of comments ensures that the dataset covers the evolvement of human conversational language. 

The GoEmotions dataset utilizes a detailed emotion classification system comprising 28 distinct labels: 27 emotions and 1 neutral category. Each emotion category was derived from the basic emotions proposed by Ekman, later considering social emotions and valenced states along with other emotions that are prevalent in digital communications. Each sample in the dataset contains 7 features/attributes: text (the Reddit comment), labels (a list of the sentiment labels: integers from 0 to 27), id (this is unique for each sample), num\_labels (number of labels), text\_length (number of  characters in the text), word\_count (number of words in the text), avg\_word\_length (average number of characters in a word).

The 28 emotion categories are \textbf{Positive Emotions} (13): admiration, amusement, approval, caring, desire, excitement, gratitude, joy, love, optimism, pride, relief, surprise. \textbf{Negative Emotions} (11): anger, annoyance, disappointment, disapproval, disgust, embarrassment, fear, grief, nervousness, remorse, sadness. \textbf{Ambiguous Emotions} (3): confusion, curiosity, realization. \textbf{Neutral} (1): neutral, lack of any emotion.

\begin{figure}
    \centering
    \includegraphics[width=0.8\textwidth]{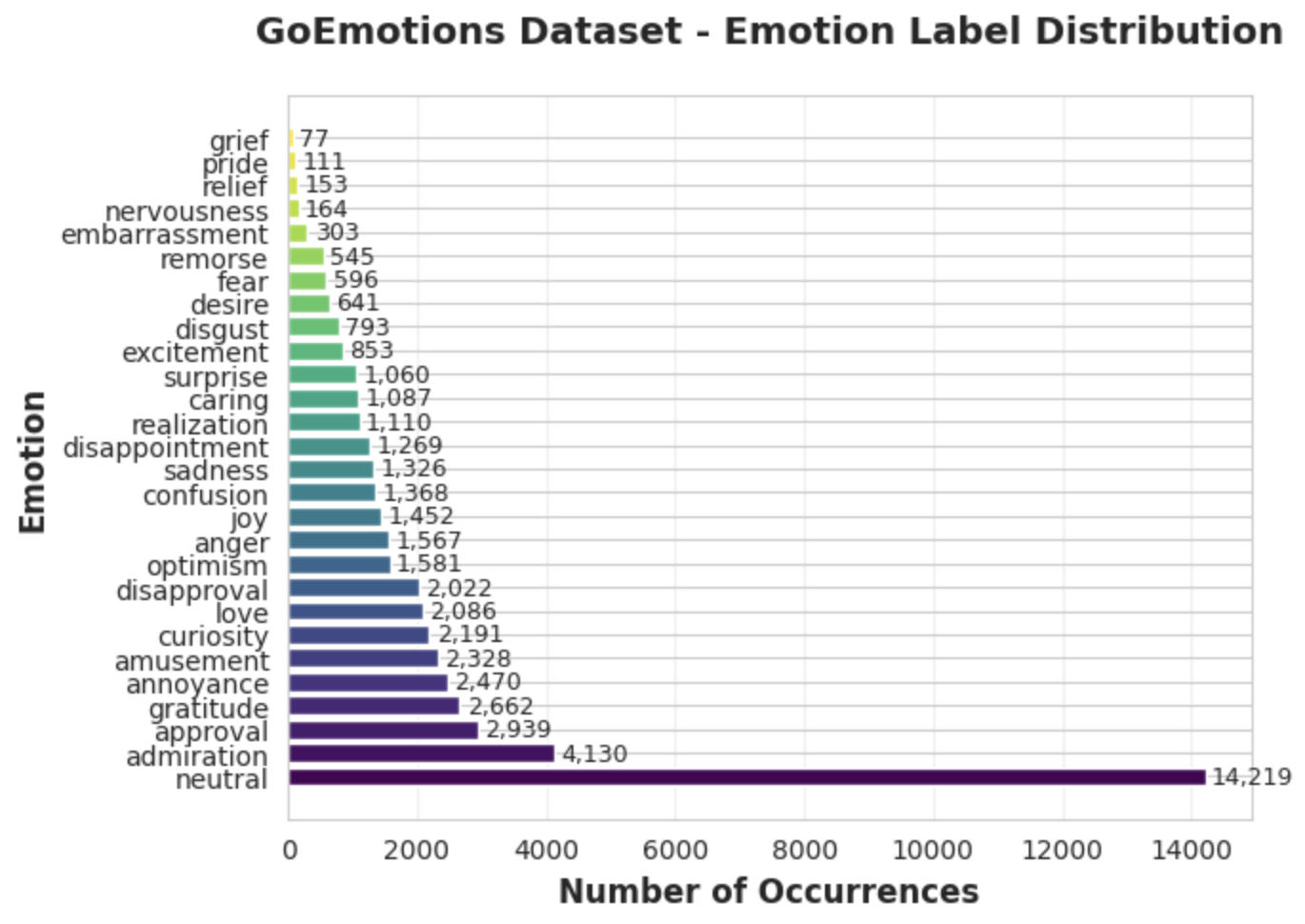}
    \caption{Emotion Label Distribution: The chart presents the distribution of emotion labels across the dataset. The most frequent emotion after neutral is is \textit{Admiration}, followed by \textit{Approval} and \textit{Gratitude}.}
    \label{fig:label_dist}
\end{figure}

The dataset consists of 58,009 samples out of which the most popular category is Neutral (14,219 samples), the second place is for the emotion Admiration (4130 samples), and the third one is approval (2,939 samples).  Figure~\ref{fig:label_dist} shows the full label distribution of the GoEmotion dataset. 

Table \ref{tab:comment_stats} shows the sequence length statistics in our dataset, giving some idea of what models will see during the training.
\begin{table}
\centering
\begin{tabular}{|l|c|}
\hline
\textbf{Metric} & \textbf{Value} \\
\hline
Average text length & 68.40 characters \\
\hline
Average word count & 12.84 words per comment \\
\hline
Median word count & 12 words \\
\hline
Comment length range & 2-151 characters (truncated for extremely long comments) \\
\hline
\end{tabular}
\caption{Comment statistics summary}
\label{tab:comment_stats}
\end{table}
Having 27 emotions documented, one cannot ensure mutual exclusivity; therefore, there are comments that have more than just one label, more precisely, the distribution of number of labels is shown in the Table \ref{tab:label_distribution}. As written, more than 80\% samples have only one label, close to 15\% of words have 2 labels and the other options of multilabeling sum up to less than 1.5\%. 
\begin{table}[h]
\centering
\begin{tabular}{|c|r|r|}
\hline
\textbf{Number of Labels} & \textbf{Samples} & \textbf{Percentage} \\
\hline
1 & 45,446 & 83.75\% \\
\hline
2 & 8,124 & 14.97\% \\
\hline
3 & 655 & 1.21\% \\
\hline
4 & 37 & 0.07\% \\
\hline
5 & 1 & 0.00\% \\
\hline
\end{tabular}
\caption{Distribution of label counts per sample}
\label{tab:label_distribution}
\end{table}

\begin{figure}
    \centering
    \includegraphics[width=0.9\textwidth]{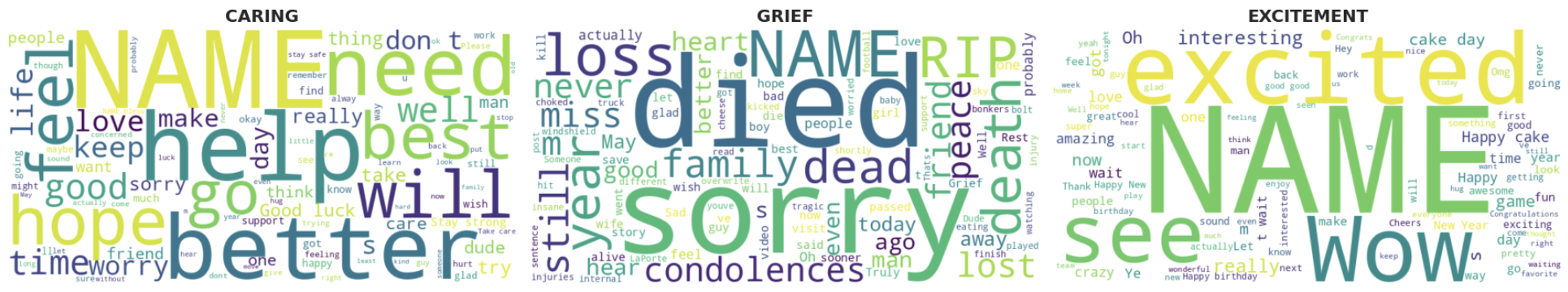}
    \caption{Wordclouds for the Emotion Categories - \textit{Caring}, \textit{Grief}, \textit{Excitement}. The token "NAME" represents anonymized user mentions in the original comments.}
    \label{fig:wordcloudname}
\end{figure}

It is important to understand how the words influence the emotion category; therefore, Figure~\ref{fig:wordcloudname} shows WordCloud plots for three very different categories: Caring, Grief, Excitement. It can be clearly seen that the most frequent words are strong indicators of the emotion they conveyed. In the first plot the words "help", "hope", "better", "will", "need" indicate support, caring, and hope that things will get better. The second WordCloud shows words that are screaming Grief, those are "loss", "died", "sorry", "RIP", "never", "condolences" etc. The third WordCloud plot gives happiness, surprise, and excitement with words such as "WOW", "excited", "amazing", "crazy", "interesting". The word "NAME" is so frequent because the GoEmotions dataset replaced usernames and personal names with special tokens [NAME] and [RELIGION] to protect user privacy.

Although the GoEmotions dataset is high quality and detailed enough for sentiment classification research, it has some key challenges that should be addressed and taken into account when choosing the models. One of them is the severe imbalance between the classes: Neutral class is the dominant one with 14k samples and after it the number of observations per class drastically decreases, and against all this, there are classes that have only 100 observations. The next critical factor is the multiclass labeling which makes things complicated; at each step, multiple emotions can be chosen. Last but not least, a major difficulty is the fine-grained nature of the dataset, the model should choose and differentiate between very similar emotions such as sadness and grief or surprise, excitement and amusement. One noteworthy thing is that the dataset represents mainly english speaking people's comments and can include their culture specific prejudices and perspectives which may not reflect people's emotional characteristics  globally.

\subsection{Proposed Framework}
\begin{figure}
    \centering
    \includegraphics[width=0.75\textwidth]{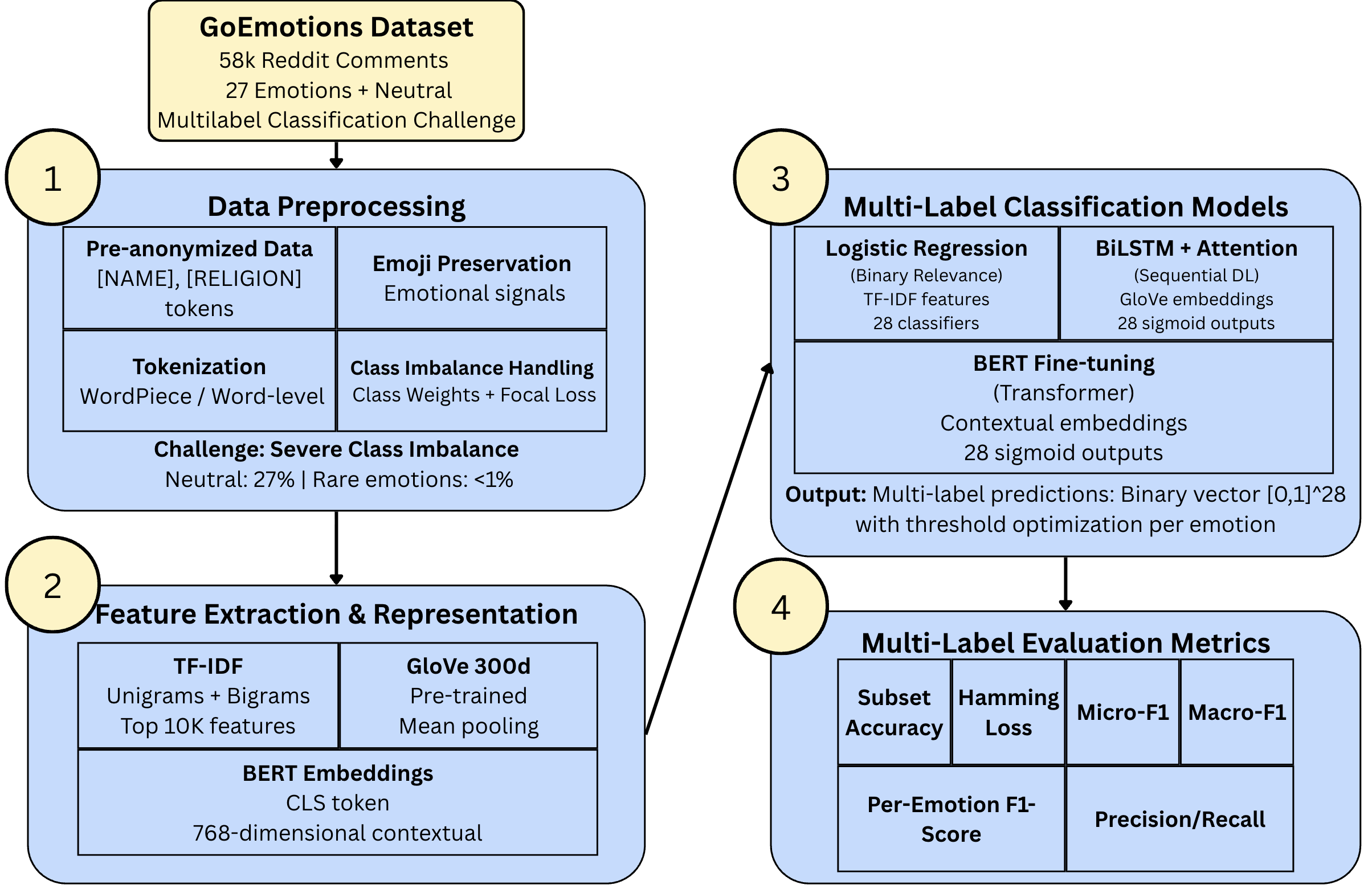}
    \caption{Pipeline Stages: We propose a 4 stage pipeline, which is adapted to the dataset's unique nature and addresses its limitations}
    \label{fig:pipeline}
\end{figure}

This paper's proposed methodology addresses the unique characteristics and the limitations of this dataset discussed earlier. The pipeline consists of 4 stages - Data Processing and Imbalance handling, Feature Extraction, Multilabel Classification Modeling and Evaluation. Figure~\ref{fig:pipeline} shows the detailed pipeline visually.

\textbf{Data Processing:}
Data Processing includes lower-casing, emoji preservation (since emojies sometimes convey 90\% of the emotion of the message), stop word preservation (because models like BERT and BiLSTM need to see the whole context to understand the true meaning) and punctuation preservation. As for the tokenization, for the classic ML model we will use word level tokenization, whereas for the neural networks like BERT we will use WordPiece tokenizer - subword tokenization. And since the dataset contains [NAME] and [RELEGION] words which are just replacements for real data, we will add them as special tokens to our vocabulary, so they will not be splitted.

\textbf{Class Imbalance Handling:} 
To handle the class imbalance, two classic strategies will be applied. During the training stage, inverse frequency weights will be assigned to the classes, which are calculated with $w_c = \frac{N}{n_c \cdot K}$, where $n_c$ is the number of samples belonging to the class c, $N$ is the number of samples and $K$ is the number of classes.
 The second method will be Focal loss.

\begin{equation}
\mathrm{FL}(p_t) = -\alpha_t (1 - p_t)^{\gamma} \log(p_t)
\end{equation}

where $p_{t}$ is the model's estimated probability for the correct class, $\alpha_t$ is the class weight, and $\gamma$ is the focusing parameter that reduces loss for well-classified examples.

\textbf{Feature Extraction:}
For the classical ML model Logistic Regression, TF-IDF will be used with unigrams and bigrams.

\begin{equation}
\mathrm{TF\!-\!IDF}(t,d,D)
=
\frac{f_{t,d}}{\sum_{t' \in d} f_{t',d}}
\cdot
\log\!\left(
\frac{1 + \lvert D \rvert}
{\lvert \{\, d \in D : t \in d \,\} \rvert}
\right)
\end{equation}

 Where $f_{t,d}$ is the frequency (or presence) of emotion t in comment $d$. $D$ is all the comments. For the BiLSTM+Attention model we will use GloVe's pretrained 300 dimensional word embeddings \cite{r7}.
\begin{equation}
\label{eq:vdoc}
    \mathbf{v}_{\text{doc}} = \frac{1}{n} \sum_{i=1}^{n} \mathbf{v}_{w_i}
\end{equation}
Equation \eqref{eq:vdoc} shows how each comment's embedding will be calculated - by taking the mean over the embeddings for the words that are in that comment.
As for the third model BERT, we will use contextual embeddings.

\begin{equation}
\mathbf{H} = \text{BERT}(X) \in \mathbb{R}^{n \times 768}
\end{equation}

Where $X$ is the input (comment) sequence tokens and the $\mathbf{H}$ is the output embeddings, $n$ is the number of tokens in that comment. After this we take \(\mathbf{h}_{\text{CLS}} = \mathbf{H}[0] \in \mathbb{R}^{768}\)
because it contains like a summary embedding, which will be later fed to the classifier model.

\textbf{Multi-Label Classification:}
For the classification part three models of different complexity will be used. The most basic one will be Logistic Regression with Binary Relevance. This will be done by breaking down into 28 independent binary classification problems.
\begin{equation}
P(y_k = 1 \mid X) = \frac{1}{1 + e^{-(\mathbf{w}_k^T \mathbf{x} + b_k)}}
\end{equation}
It will give 28 different independent probabilities for the comment $X$ belonging to the class $y_k$.

Bidirectional LSTM with Attention mechanism
LSTM will be implemented - a reasonable choice for sequential data, bidirectionality will help to see the text from both directions and the attention block will capture the words conveying the most emotion. It will return independent probabilities from 0 to 1 for each emotion, allowing to have multiple labels per comment. The detailed mathematical formulations are shown below. Equation \eqref{eq:lstm} shows how the usual LSTM block works, \eqref{eq:bi} adds the bidirectionality - two LSTM blocks with different directionalities, \eqref{eq:att} constructs the attention block on top of BiLSTM and finally equation \eqref{eq:out} gives the output probability vector, indicating whether each emotion label is present or not in the given sample.

\textbf{LSTM Block}
\begin{equation}
\label{eq:lstm}
\begin{aligned}
\mathbf{f}_t &= \sigma(\mathbf{W}_f [\mathbf{h}_{t-1}, \mathbf{x}_t] + \mathbf{b}_f)\\
\mathbf{i}_t &= \sigma(\mathbf{W}_i [\mathbf{h}_{t-1}, \mathbf{x}_t] + \mathbf{b}_i)\\
\tilde{\mathbf{C}}_t &= \tanh(\mathbf{W}_C [\mathbf{h}_{t-1}, \mathbf{x}_t] + \mathbf{b}_C)\\
\mathbf{C}_t &= \mathbf{f}_t \odot \mathbf{C}_{t-1} + \mathbf{i}_t \odot \tilde{\mathbf{C}}_t\\
\mathbf{o}_t &= \sigma(\mathbf{W}_o [\mathbf{h}_{t-1}, \mathbf{x}_t] + \mathbf{b}_o)\\
\mathbf{h}_t &= \mathbf{o}_t \odot \tanh(\mathbf{C}_t)
\end{aligned}
\end{equation}

\textbf{Bidirectional nature}
\begin{equation}
\label{eq:bi}
\begin{aligned}
\overrightarrow{\mathbf{h}}_t &= \text{LSTM}_{\text{forward}}(\mathbf{e}_t, \overrightarrow{\mathbf{h}}_{t-1})\\
\overleftarrow{\mathbf{h}}_t &= \text{LSTM}_{\text{backward}}(\mathbf{e}_t, \overleftarrow{\mathbf{h}}_{t+1})\\
\mathbf{h}_t &= [\overrightarrow{\mathbf{h}}_t; \overleftarrow{\mathbf{h}}_t]
\end{aligned}
\end{equation}

\textbf{Attention block}
\begin{equation}
\label{eq:att}
\alpha_t = \frac{\exp(\mathbf{h}_t^T \mathbf{W}_a \mathbf{h}_t)}{\sum_{j=1}^{n} \exp(\mathbf{h}_j^T \mathbf{W}_a \mathbf{h}_j)}, \quad
\mathbf{c} = \sum_{t=1}^{n} \alpha_t \mathbf{h}_t
\end{equation}

\textbf{Multi-label output}
\begin{equation}
\label{eq:out}
\hat{\mathbf{y}} = \sigma(\mathbf{W}_o \mathbf{c} + \mathbf{b}_o) \in [0,1]^{28}
\end{equation}

BERT FineTuning: For the third model, we will take BERT and finetune it with a multilabel head, from BERT we will get the contextual embeddings and then we will put a layer on usual BERT which will map it to 28 dimensional space (28 probabilities) allowing to have multiple emotions. Firstly, we will freeze all the layers updating only the final layer's weights, then we will unfreeze all the layers and let the model learn. Binary Cross Entropy loss combined with Focal Loss will be used  for the loss function. The detailed mathematical equations are shown below.

\textbf{CLS representation}
\begin{equation}
\label{eq:hcls}
\mathbf{h}_{\text{CLS}} = \text{BERT}(X)[0]
\end{equation}

\textbf{Multi-label classification head}
\begin{equation}
\label{eq:zz}
\mathbf{z} = \mathbf{W} \mathbf{h}_{\text{CLS}} + \mathbf{b}, \quad
\hat{\mathbf{y}} = \sigma(\mathbf{z}) \in [0,1]^{28}
\end{equation}

\textbf{Binary Cross-Entropy Loss}
\begin{equation}
\label{eq:bce}
\mathcal{L} = -\frac{1}{N \cdot K} \sum_{i=1}^{N} \sum_{k=1}^{K} \Big[ y_{i,k} \log(\hat{y}_{i,k}) + (1 - y_{i,k}) \log(1 - \hat{y}_{i,k}) \Big]
\end{equation}

\textbf{Focal Loss}
\begin{equation}
\label{eq:focal}
\mathcal{L}_{\text{focal}} = -\frac{1}{N \cdot K} \sum_{i=1}^{N} \sum_{k=1}^{K} \alpha_k (1 - \hat{y}_{i,k})^\gamma 
\Big[ y_{i,k} \log(\hat{y}_{i,k}) + (1 - y_{i,k}) \log(1 - \hat{y}_{i,k}) \Big]
\end{equation}

Equation \eqref{eq:hcls} shows the calculation of single summarizing vector which represents the whole sentence. \eqref{eq:zz} gives the final output vector with the predicted labels. BCE loss \eqref{eq:bce} is the standard loss function for multi-label classification problems. It penalizes when the model assigns low probability for the present emotion or high probability for the absent emotion. Focal loss \eqref{eq:focal} is added to the standard loss function to address the severe class imbalance: it is done to focus the model more on misclassified examples. It automatically assigns lower weights to the easy examples when the model is very confident and higher weights to the difficult ones, helping the model to learn valuable knowledge.

\subsection{Evaluation Measures/Metrics}
Different types of metrics will be used to evaluate the performance of our models, more specifically, Subset Accuracy, Hamming Loss. The formulas are shown below. Subset accuracy - as shown in equation \eqref{eq:subset} counts all the occurrences when the whole set of labels was predicted correctly. For each sample, if the predicted set is correct, then it is increased by 1/N. The method that is used by Hamming Loss in equation \eqref{eq:hamming} is less strict: it considers every pair of sample and label, and if the prediction is incorrect add 1 and later divide the number by $N \cdot K$ to result in a number from 0 to 1. 
\begin{equation}
\label{eq:subset}
\text{Subset Accuracy} = \frac{1}{N} \sum_{i=1}^{N} \mathbf{1} \big( \hat{Y}_i = Y_i \big)
\end{equation}
\begin{equation}
\label{eq:hamming}
\text{Hamming Loss} = \frac{1}{N \cdot K} \sum_{i=1}^{N} \sum_{k=1}^{K} \mathbf{1} \big( \hat{y}_{i,k} \neq y_{i,k} \big)
\end{equation}

In addition to this, per emotion analysis will be done where Precision, Recall and F1 will be calculated for each emotion category   \eqref{eq:labelmet}. Precision calculates how often the model was right when predicting certain emotion. Recall calculates out of all real cases of specific emotion, how often the model was right by predicting that same emotion. F1 is trying to balance both precision and recall by taking their harmonic mean and thus giving weight to both of them. In our paper, we report two types of F1 score per model: macro F1 and micro F1. Macro F1 is the fair metric because it calculates individual F1 scores for each emotion label and then averages them, treating all the labels equally, while the Micro F1 calculates True Positives, False Positives and False Negatives over all emotions at the beginning and then, calculates the F1 score with TP, FP and FN. It is significantly affected by the frequent emotions, so failing in rare emotions will not be reflected in the Micro F1 score, instead, Macro F1 score will be influenced.
\begin{equation}
\label{eq:labelmet}
\text{Precision}_k = \frac{TP_k}{TP_k + FP_k}, \quad
\text{Recall}_k = \frac{TP_k}{TP_k + FN_k}, \quad
\text{F1}_k = \frac{2 \cdot \text{Precision}_k \cdot \text{Recall}_k}{\text{Precision}_k + \text{Recall}_k}
\end{equation}

\section{Experiments, Results and Discussion}
All three models were trained on the official train-val-test split of GoEmotions Dataset. The first model - logistic regression with binary relevance was trained with liblinear solver, with max iterations=1000 and class weight set balanced. For the BiLSTM model, we used the following hyperparameter setup, Embedding dim=300, Hidden size=256, Layers=2 (bidirectional), Max sequence length=128, Dropout=0.3, Batch size=64, Optimizer=Adam(lr=1e-3). The model was trained in total of 9 epochs. BERT was finetuned with the following configuration - Max length=64, Batch size=16, Learning rate=2e-5, Optimizer=AdamW. we trained the model for 1 epoch only the head (the encoder was frozen) and then 4 epochs of full finetuning. The results were both expected and surprising. Logistic Regression gave a strong baseline for this dataset based on the TF-IDF features and explicit and strong lexical cues. BiLSTM+Attention seemed to struggle a little bit for a number of reasons. Since it was trained on GloVe embeddings, which were static and no context was included, it made the process harder. Not to mention, it is a known thing that neural networks are data hungry models that require a lot of data not to overfit. The last model - the BERT performed pretty good as expected, because it used contextual embedding vectors and also BERT is famous for its ability to understand semantic meaning within the text. 

\begin{figure}
    \centering
    \includegraphics[width=0.8\textwidth]{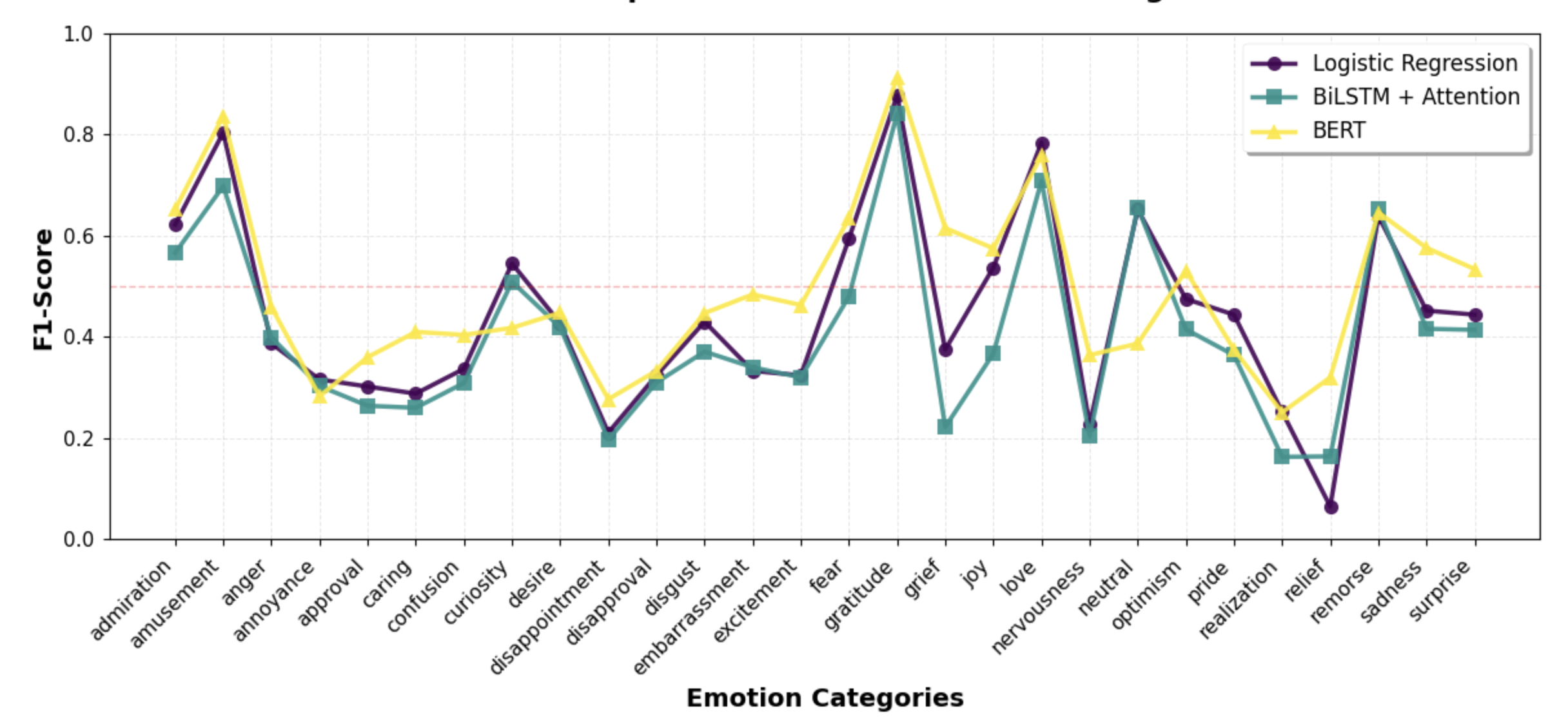}
    \caption{F1 Score comparison across all emotion categories}
    \label{fig:f1_scores_comparison}
\end{figure}

\begin{figure}
    \centering
    \includegraphics[width=0.8\textwidth]{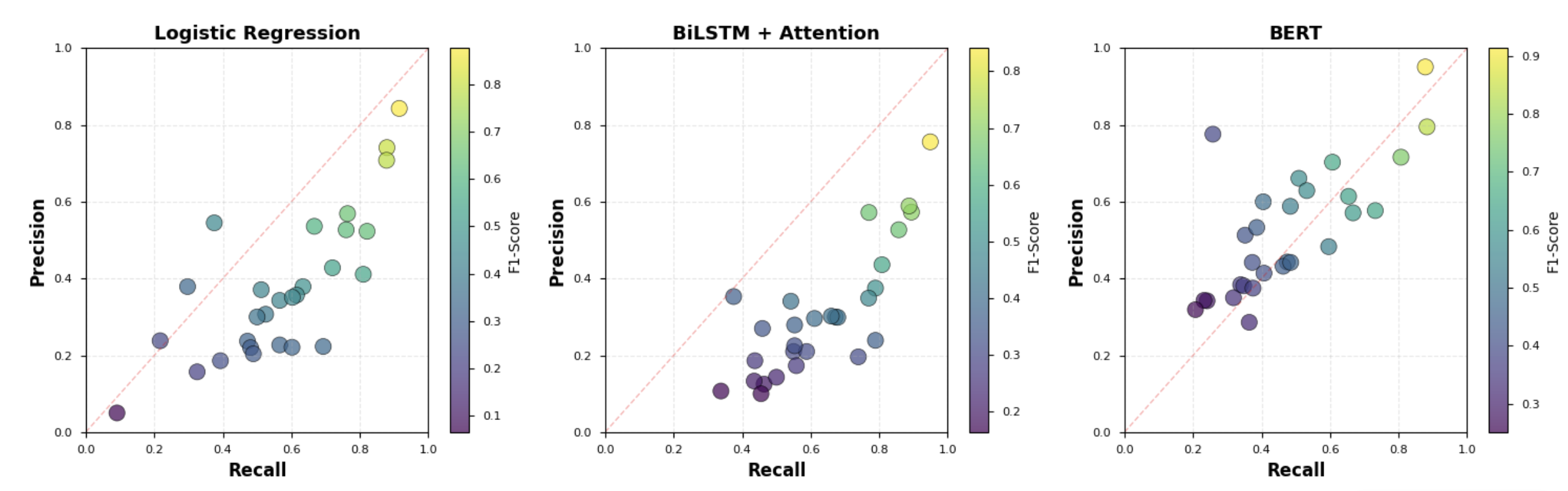}
    \caption{Precision-Recall Trade off analysis for each model}
    \label{fig:recall_precision}
\end{figure}


\begin{table}[ht]
\centering
\caption{Logistic Regression classification report}
\label{tab:logreg_report_split}

\begin{minipage}[t]{0.49\textwidth}
\centering
\begin{tabular}{lcccc}
\toprule
Label & Prec. & Rec. & F1 & Sup. \\
\midrule
admiration      & 0.527 & 0.760 & 0.622 & 504 \\
amusement       & 0.741 & 0.879 & 0.804 & 264 \\
anger           & 0.307 & 0.525 & 0.387 & 198 \\
annoyance       & 0.237 & 0.472 & 0.316 & 320 \\
approval        & 0.220 & 0.481 & 0.302 & 351 \\
caring          & 0.204 & 0.489 & 0.288 & 135 \\
confusion       & 0.223 & 0.693 & 0.337 & 153 \\
curiosity       & 0.411 & 0.810 & 0.545 & 284 \\
desire          & 0.343 & 0.566 & 0.427 & 83  \\
disappointment  & 0.157 & 0.325 & 0.211 & 151 \\
disapproval     & 0.227 & 0.566 & 0.324 & 267 \\
disgust         & 0.371 & 0.512 & 0.430 & 123 \\
embarrassment   & 0.379 & 0.297 & 0.333 & 37  \\
excitement      & 0.221 & 0.602 & 0.324 & 103 \\
\bottomrule
\end{tabular}
\end{minipage}
\hfill
\begin{minipage}[t]{0.49\textwidth}
\centering
\begin{tabular}{lcccc}
\toprule
Label & Prec. & Rec. & F1 & Sup. \\
\midrule
fear            & 0.536 & 0.667 & 0.594 & 78  \\
gratitude       & 0.843 & 0.915 & 0.877 & 352 \\
grief           & 0.300 & 0.500 & 0.375 & 6   \\
joy             & 0.428 & 0.720 & 0.537 & 161 \\
love            & 0.708 & 0.878 & 0.784 & 238 \\
nervousness     & 0.238 & 0.217 & 0.227 & 23  \\
optimism        & 0.379 & 0.634 & 0.475 & 186 \\
pride           & 0.545 & 0.375 & 0.444 & 16  \\
realization     & 0.186 & 0.393 & 0.252 & 145 \\
relief          & 0.050 & 0.091 & 0.065 & 11  \\
remorse         & 0.523 & 0.821 & 0.639 & 56  \\
sadness         & 0.357 & 0.615 & 0.452 & 156 \\
surprise        & 0.351 & 0.603 & 0.444 & 141 \\
neutral         & 0.569 & 0.764 & 0.652 & 1787 \\
\bottomrule
\end{tabular}
\end{minipage}

\end{table}


\begin{table}[ht]
\centering
\caption{BiLSTM + Attention classification report}
\label{tab:bilstm_attention_report_split}

\begin{minipage}[t]{0.49\textwidth}
\centering
\begin{tabular}{lcccc}
\toprule
Label & Prec. & Rec. & F1 & Sup. \\
\midrule
admiration      & 0.436 & 0.808 & 0.566 & 504 \\
amusement       & 0.573 & 0.894 & 0.698 & 264 \\
anger           & 0.296 & 0.611 & 0.399 & 198 \\
annoyance       & 0.210 & 0.550 & 0.303 & 320 \\
approval        & 0.173 & 0.558 & 0.264 & 351 \\
caring          & 0.186 & 0.437 & 0.260 & 135 \\
confusion       & 0.196 & 0.739 & 0.309 & 153 \\
curiosity       & 0.375 & 0.789 & 0.509 & 284 \\
desire          & 0.341 & 0.542 & 0.419 & 83  \\
disappointment  & 0.125 & 0.464 & 0.197 & 151 \\
disapproval     & 0.210 & 0.588 & 0.310 & 267 \\
disgust         & 0.279 & 0.553 & 0.371 & 123 \\
embarrassment   & 0.270 & 0.459 & 0.340 & 37  \\
excitement      & 0.225 & 0.553 & 0.320 & 103 \\
\bottomrule
\end{tabular}
\end{minipage}
\hfill
\begin{minipage}[t]{0.49\textwidth}
\centering
\begin{tabular}{lcccc}
\toprule
Label & Prec. & Rec. & F1 & Sup. \\
\midrule
fear            & 0.349 & 0.769 & 0.480 & 78  \\
gratitude       & 0.756 & 0.949 & 0.841 & 352 \\
grief           & 0.143 & 0.500 & 0.222 & 6   \\
joy             & 0.239 & 0.789 & 0.367 & 161 \\
love            & 0.589 & 0.887 & 0.708 & 238 \\
nervousness     & 0.133 & 0.435 & 0.204 & 23  \\
optimism        & 0.300 & 0.672 & 0.415 & 186 \\
pride           & 0.353 & 0.375 & 0.364 & 16  \\
realization     & 0.107 & 0.338 & 0.163 & 145 \\
relief          & 0.100 & 0.455 & 0.164 & 11  \\
remorse         & 0.527 & 0.857 & 0.653 & 56  \\
sadness         & 0.299 & 0.679 & 0.416 & 156 \\
surprise        & 0.302 & 0.660 & 0.414 & 141 \\
neutral         & 0.572 & 0.770 & 0.656 & 1787 \\
\bottomrule
\end{tabular}
\end{minipage}

\end{table}


\begin{table}[ht]
\centering
\caption{BERT classification report}
\label{tab:bert_report_split}

\begin{minipage}[t]{0.49\textwidth}
\centering
\begin{tabular}{lcccc}
\toprule
Label & Prec. & Rec. & F1 & Sup. \\
\midrule
admiration      & 0.703 & 0.607 & 0.652 & 504 \\
amusement       & 0.795 & 0.883 & 0.837 & 264 \\
anger           & 0.443 & 0.475 & 0.459 & 198 \\
annoyance       & 0.342 & 0.241 & 0.283 & 320 \\
approval        & 0.384 & 0.339 & 0.360 & 351 \\
caring          & 0.414 & 0.407 & 0.410 & 135 \\
confusion       & 0.442 & 0.373 & 0.404 & 153 \\
curiosity       & 0.513 & 0.352 & 0.418 & 284 \\
desire          & 0.533 & 0.386 & 0.448 & 83  \\
disappointment  & 0.343 & 0.232 & 0.277 & 151 \\
disapproval     & 0.350 & 0.318 & 0.333 & 267 \\
disgust         & 0.432 & 0.463 & 0.447 & 123 \\
embarrassment   & 0.600 & 0.405 & 0.484 & 37  \\
excitement      & 0.442 & 0.485 & 0.463 & 103 \\
\bottomrule
\end{tabular}
\end{minipage}
\hfill
\begin{minipage}[t]{0.49\textwidth}
\centering
\begin{tabular}{lcccc}
\toprule
Label & Prec. & Rec. & F1 & Sup. \\
\midrule
fear            & 0.614 & 0.654 & 0.634 & 78  \\
gratitude       & 0.951 & 0.878 & 0.913 & 352 \\
grief           & 0.571 & 0.667 & 0.615 & 6   \\
joy             & 0.661 & 0.509 & 0.575 & 161 \\
love            & 0.716 & 0.807 & 0.759 & 238 \\
nervousness     & 0.381 & 0.348 & 0.364 & 23  \\
optimism        & 0.588 & 0.484 & 0.531 & 186 \\
pride           & 0.375 & 0.375 & 0.375 & 16  \\
realization     & 0.319 & 0.207 & 0.251 & 145 \\
relief          & 0.286 & 0.364 & 0.320 & 11  \\
remorse         & 0.577 & 0.732 & 0.646 & 56  \\
sadness         & 0.629 & 0.532 & 0.576 & 156 \\
surprise        & 0.483 & 0.596 & 0.534 & 141 \\
neutral         & 0.776 & 0.258 & 0.387 & 1787 \\
\bottomrule
\end{tabular}
\end{minipage}

\end{table}

\begin{table}
\centering
\captionsetup{justification=centering}
\caption{Results: Model Comparison}
\label{tab:model_comparison}

\begin{tabular}{lcccc}
\toprule
Model & Subset Accuracy & Micro F1 & Macro F1 & Hamming Loss \\
\midrule
\textbf{Logistic Regression} & \textasciitilde 0.24 & \textbf{\textasciitilde 0.51} & \textasciitilde 0.45 & \textasciitilde 0.054 \\
\textbf{BiLSTM + Attention}  & \textasciitilde 0.17 & \textasciitilde 0.48 & \textasciitilde 0.43 & \textasciitilde 0.039 \\
\textbf{BERT}                & \textbf{\textasciitilde 0.36} & \textasciitilde 0.50 & \textbf{\textasciitilde 0.49} & \textbf{\textasciitilde 0.036} \\
\bottomrule
\end{tabular}
\end{table}

Table~\ref{tab:logreg_report_split}, Table~\ref{tab:bilstm_attention_report_split} and Table~\ref{tab:bert_report_split} report detailed comprehensive metrics for each model performance. Figure~\ref{fig:f1_scores_comparison} integrates all the F1 scores for the models into one graph giving an explicit comparison of the three models. For the majority of the emotions, BERT has scored the highest F1 values, second place is for the logistic regression, and only in very few emotions BiLSTM with Attention had the highest score. Figure~\ref{fig:recall_precision} shows a scatter plot of Precision and Recall trade off for each of three models over all emotion types, with red lines indicating the balancing boundary between two metrics. By the coloring of dots, F1 scores are also shown in the graph. As the graph indicates, Logistic Regression and BiLSTM with Attention models struggle to score high for both metrics, when increasing Recall, Precision scores are decreased. However, BERT is showing significant improvement, by better balanced Recall and Precision scores. Not to mention, the points are concentrated more on the upper right corner compared to the other two models.

As shown in the results, one can see that Logistic Regression performed well and had high precision for frequent emotions and struggled with the rare emotions. BiLSTM model had good precision on emotional words with strong lexical signals and had lower precision on ambiguous emotions (approval, annoyance, confusion). BERT had overall the best results in terms of precision and handled rare labels better.

As shown in Table~\ref{tab:model_comparison}, Logistic Regression unexpectedly gave the highest Micro F1 score (Micro F1=0.51), which means that many emotions were just triggered with straightforward lexical cues. Whereas, for the other metrics, BERT achieved the highest values, which was expected, it had Macro F1 = 0.49, meaning that it handled rare emotions much better, Hamming Loss=0.036 and Subset Accuracy=0.36. The official paper of GoEmotions dataset \cite{r4} reports a Macro F1 score of 0.46 for BERT model, however, in our finetuning experiment, we achieve 0.49, improving this number. It is important to consider the computational resources when comparing model results and performances. Lack of availability of huge GPU resources has become a reason for not training and finetuning more powerful models, but given the existing conditions, we got reasonably good results.

\section{Conclusion, Limitations and Future Directions}
In conclusion, we trained three models of different complexity and expressiveness on GoEmotions Dataset. We started from basic Logistic Regression, continued with BiLSTM+Attention and lastly, we finetuned a BERT model. Key Findings were that BERT achieved the best overall performance, especially for rare emotions. Logistic Regression remained a strong baseline due to clear lexical cues. BiLSTM underperformed without contextual embeddings. We gave a comprehensive comparison of three model families by computing number of metrics and a clear analysis of per-emotion behavior and class imbalance effects. Some of the limitations were that neutral class performance was impacted by class weighting, BiLSTM was limited by static word GloVe embeddings and finally, computational constraints restricted deeper fine-tuning and larger models. For the Future Work we suggest threshold calibration, adding per-label threshold: A different decision threshold for each emotion has to be tuned on the validation set, instead of a fixed default of 0.5, in order to balance better precision/recall across frequent vs. rare labels. It can lift macro-F1 and subset accuracy. In addition, finding and leveraging model label dependencies (beyond Binary Relevance) will be useful - applying methods that intrinsically model emotion cooccurrence, such as classifier chains or a neural multi-label head that learns label dependencies. This will address common GoEmotions failure modes where related emotions are present together. And finally, we belive contextualizing the BiLSTM will improve its current performance - moving beyond static GloVe by using contextual embeddings either as input to the BiLSTM, or a BiLSTM-on-top-of-BERT-features setup. This directly tests whether the underperformance of the BiLSTM was due to the static word representations rather than the sequence model itself.

\section*{Informed consent} Not Applicable

\section*{Data availability} The data is available on a reasonable request to the corresponding author.

\section*{Acknowledgments} Not Applicable

\section*{Conflicts of interest} The authors declare no conflicts of interest.

\end{document}